\documentclass[10pt,twocolumn]{article}
\usepackage[letterpaper,margin=0.70in,columnsep=0.24in]{geometry}
\usepackage[T1]{fontenc}
\usepackage{lmodern}
\usepackage{microtype}
\usepackage{graphicx}
\usepackage{booktabs}
\usepackage{tabularx}
\usepackage{array}
\usepackage{amsmath,amssymb}
\usepackage{enumitem}
\usepackage{natbib}
\usepackage{xcolor}
\usepackage{xurl}
\usepackage{hyperref}
\usepackage{balance}
\usepackage{caption}
\usepackage{subcaption}
\usepackage{multirow}
\usepackage{makecell}

\definecolor{navy}{HTML}{17365D}
\definecolor{teal}{HTML}{2E8B8B}
\hypersetup{colorlinks=true,linkcolor=navy,citecolor=teal,urlcolor=navy}
\setlist[itemize]{leftmargin=*,topsep=2pt,itemsep=1pt}
\setlist[enumerate]{leftmargin=*,topsep=2pt,itemsep=1pt}
\title{\vspace{-1.2em}\textbf{From Inference Engine to Inference Control Plane:}\\
\textbf{Connecting vLLM, llm-d, and the Evolution of Efficient Distributed LLM Serving}\\[0.35em]
\large A Systems Synthesis and Research Agenda}
\author{Twinkll Sisodia\\\texttt{twinklls@bu.edu}}
\date{September 2026}

\begin{document}
\maketitle
\vspace{-1.5em}

\begin{abstract}
Large-language-model inference has changed from a problem of making one model process requests efficiently into a problem of coordinating state, phases, accelerators, and service-level objectives across a fleet. This paper connects that transition across systems research, open-source implementations, and documented production studies. The starting point is the engine layer: iteration-level scheduling in Orca, PagedAttention and continuous batching in vLLM, kernel-level attention improvements, chunked prefill, quantization, and long-context execution. The optimization boundary then moves outward. Splitwise and DistServe separate prefill from decode; Preble, Mooncake, and MemServe make reusable KV state a distributed scheduling concern; Llumnix treats request state as movable; Dynamo and llm-d add routing, cache intelligence, flow control, autoscaling, heterogeneous placement, and production resilience above model servers.

The contribution is synthesis, not a new benchmark. Reported speedups, latency reductions, energy savings, and deployment outcomes remain attributed to their original authors and organizations. The paper develops three conclusions from the combined evidence. First, vLLM and llm-d are best understood as complementary layers: the engine optimizes execution while the control plane optimizes where, when, and under what policy execution occurs. Second, the scarce resource in modern inference is shifting from raw FLOPs alone toward managed state, placement, network movement, and decision quality. Third, the next scheduler may need to choose an \emph{execution plan}, not merely an endpoint. We therefore propose an Inference Execution Planner that selects among aggregated, prefill/decode, and encode/prefill/decode topologies; cache source and tier; transfer versus recomputation; hardware variants; routing and admission policies; and slower autoscaling/capacity actions. We close with an evidence-based roadmap for reliability, multimodal and agentic serving, cost-aware control, and reproducible evaluation.
\end{abstract}


\section{Introduction}
The first generation of large-language-model (LLM) serving systems had an apparently local objective: make an expensive model answer more requests per accelerator without violating latency targets. That objective remains important, but the system boundary around it has expanded. A production request may now traverse a gateway, a cache-aware router, a prefill pool, a high-speed KV-transfer fabric, a decode pool, a multi-tier cache, an autoscaler, and observability components before the response reaches the client. Agentic workloads can return repeatedly with hundreds of thousands of reused input tokens. Vision-language models add encoders whose resource profile differs from language prefill and decode. Mixture-of-experts (MoE) models introduce another dimension of parallelism and communication. Heterogeneous fleets make ``one replica equals one unit of capacity'' an increasingly weak abstraction.

This evolution is visible in the research record. Orca introduced iteration-level scheduling for autoregressive generation and reported a 36.9$\times$ throughput improvement over FasterTransformer at comparable latency for its GPT-3 175B evaluation \citep{yu2022orca}. vLLM then attacked a different bottleneck: fragmentation and duplication of the KV cache. PagedAttention applies operating-system-style paging to attention state, enabling larger effective batches and reporting 2--4$\times$ throughput over prior serving baselines at similar latency \citep{kwon2023pagedattention}. The engine has continued to evolve: vLLM V1 reported up to 1.7$\times$ higher throughput than V0 in its project benchmarks while reducing CPU scheduling overhead \citep{vllm2025v1}, and the current project supports continuous batching, chunked prefill, prefix caching, quantization, speculative decoding, multiple parallelism strategies, and disaggregated encode/prefill/decode mechanisms \citep{vllm2026repo}.

At the same time, the research community began treating inference as a distributed-state and cluster-control problem. Sarathi-Serve demonstrated the value of chunking long prefills to reduce interference with decode \citep{agrawal2024sarathi}. Splitwise and DistServe separated prefill and decode because the two phases stress hardware differently and have different user-visible latency objectives \citep{patel2024splitwise,zhong2024distserve}. Preble co-optimized prefix reuse and distributed load \citep{srivatsa2024preble}; Mooncake and MemServe elevated KV state into a distributed memory hierarchy \citep{qin2024mooncake,hu2024memserve}; and Llumnix showed that live request-state migration can be used to rebalance running work \citep{sun2024llumnix}. NVIDIA Dynamo and llm-d now expose many of these ideas as control-plane mechanisms above engines rather than as isolated model-server features \citep{nvidia2026dynamo,llmd2026v08}.

The resulting architectural question is more consequential than a comparison of two open-source projects. \emph{What should the optimization boundary of an inference system be?} If a model server can execute a request efficiently but the wrong endpoint is selected, a useful prefix can be recomputed. If prefill and decode are split without accounting for network transfer, TTFT can get worse. If an autoscaler reacts to GPU utilization without understanding queueing and cache warmness, adding capacity can temporarily reduce locality. If routing chases cache locality without a saturation escape valve, hot replicas become queueing hotspots. These are system interactions, not isolated kernel problems.

This paper develops an evidence-grounded view of that transition. It argues that vLLM and llm-d should not be framed as substitutes. vLLM is an execution engine and serving runtime; llm-d is increasingly an engine-agnostic inference control plane that composes routing, cache-state awareness, disaggregation, flow control, autoscaling, and reliability around engines such as vLLM and SGLang \citep{llmd2026v08,llmd2026v09}. NVIDIA Dynamo independently reflects a similar architectural split, strengthening the interpretation that the field is converging on a control layer around increasingly capable engines \citep{nvidia2026dynamo}.

\subsection{Claim and attribution boundary}
This is a systems synthesis and research agenda, not a new experimental performance paper. No benchmark result reported here is an original measurement. Numerical gains belong to the cited systems under their own hardware, models, request distributions, software versions, and baselines. In particular, the llm-d evidence atlas in Sections~\ref{sec:evidence} and~\ref{sec:benchmarks} reproduces or computes ratios from official project, cloud-provider, and production reports; those numbers are \textbf{not} a common benchmark and must not be averaged into a single performance score.

The original contribution is the connection among the evidence: (i) a taxonomy of how the optimization boundary has expanded; (ii) the distinction between engine-local execution and fleet-level inference control; (iii) a bottleneck-migration model centered on state, placement, and decision quality; (iv) the proposed \emph{Inference Execution Planner} (IEP); and (v) a research and industry roadmap for evaluating that control loop safely and reproducibly.

\section{Scope and Evidence Method}
The review prioritizes work from 2022 through September 13, 2026. Sources are deliberately separated by evidence type because a peer-reviewed systems paper, an arXiv preprint, an open-source release, and a vendor benchmark answer different questions.

\begin{table*}[t]
\centering
\small
\caption{Evidence hierarchy used in this synthesis. The hierarchy describes evidence type, not a universal ranking of scientific quality.}
\label{tab:evidence}
\begin{tabularx}{\textwidth}{p{0.08\textwidth} p{0.19\textwidth} X X}
\toprule
\textbf{Tier} & \textbf{Evidence} & \textbf{Representative sources} & \textbf{How it is used} \\
\midrule
A & Peer-reviewed systems research & Orca; PagedAttention/vLLM; Sarathi-Serve; Splitwise; DistServe; Llumnix; FlashAttention-3; RaidServe & Strongest basis for architectural mechanisms and controlled experimental findings. \\
B & Academic preprints / research prototypes & Mooncake; Preble; MemServe; MInference; Mnemosyne; M*; energy studies & Frontier mechanisms and hypotheses; results remain conditioned on their evaluation settings. \\
C & Open-source implementation and project evidence & vLLM repository and docs; llm-d releases, routing/KV/autoscaling/tracing studies; NVIDIA Dynamo docs & Establishes current capabilities and implementation direction; benchmark results are project-specific. \\
D & Production / cloud-provider reports & Google Vertex routing observations; Tesla/Red Hat/KServe report; AWS and OCI llm-d studies & Evidence of production feasibility and operator constraints; outcomes are organization-reported, not independent causal estimates. \\
\bottomrule
\end{tabularx}
\end{table*}

The paper does not attempt a PRISMA-style exhaustive systematic review. Selection favors work that changes at least one of five serving decisions: \emph{execution}, \emph{state management}, \emph{placement}, \emph{capacity}, or \emph{reliability}. Training-only optimization is outside scope. Sustainability appears when it changes inference control decisions, but the paper intentionally does not repeat a sustainability-centered architecture; the primary question here is the evolution of the serving-system boundary.

A second methodological choice concerns benchmark comparison. Cross-paper speedup curves are often misleading because inference results depend on model architecture, precision, context length, output length, batch/concurrency, accelerator, interconnect, SLO, cache-hit distribution, and baseline tuning. Accordingly, this paper uses two kinds of quantitative visualization. First, it reports source-local comparisons exactly where the source defines the baseline. Second, it uses an ``evidence atlas'' that places such ratios next to one another while explicitly marking them as non-comparable. The atlas is evidence that multiple control-plane mechanisms expose meaningful headroom; it is not evidence that one mechanism is universally best.

\section{How the Optimization Boundary Expanded}
\begin{figure*}[t]
\centering
\includegraphics[width=0.98\textwidth]{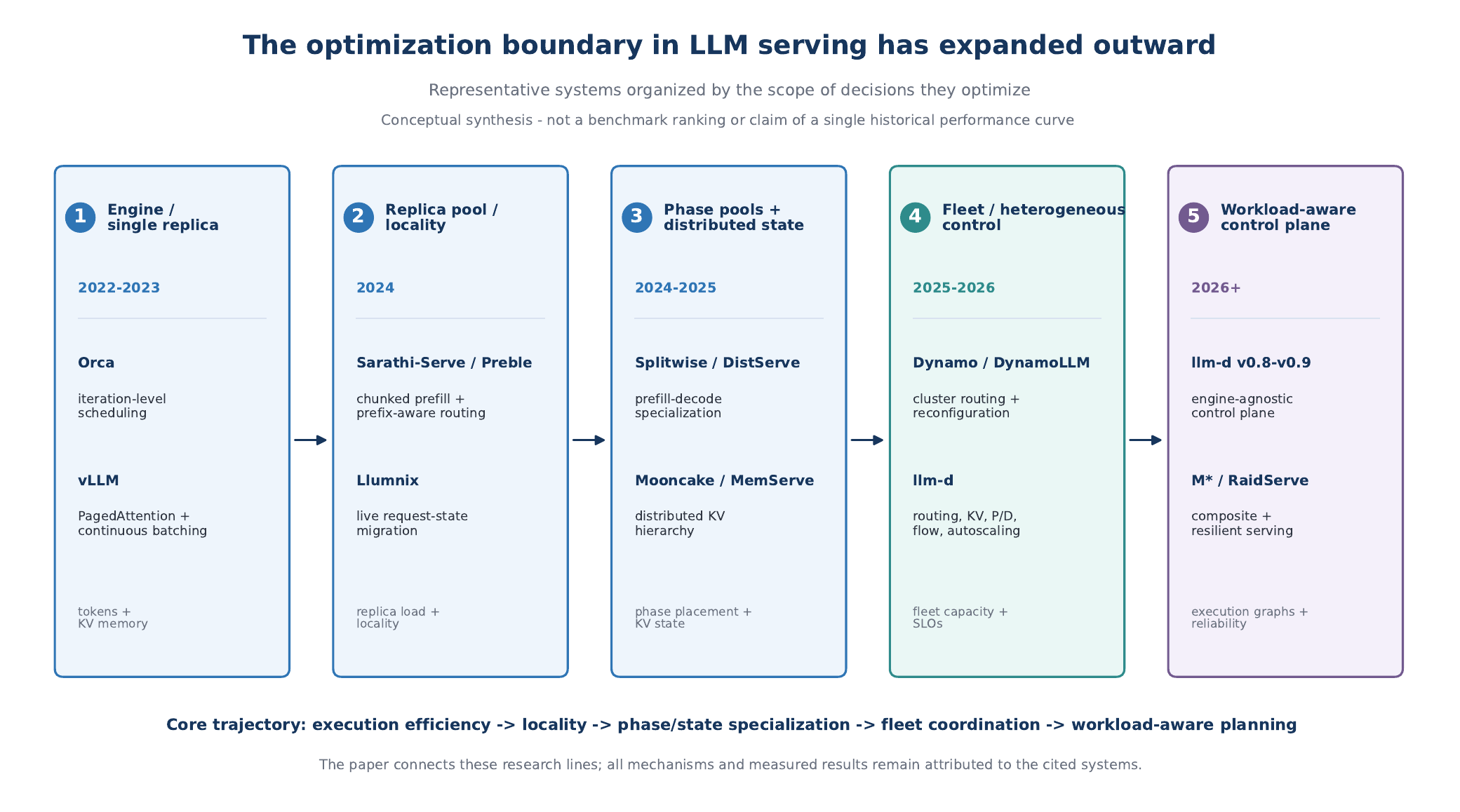}
\caption{Representative systems placed by the scope of decisions they optimize. The trajectory is conceptual rather than a performance curve. Research at lower layers remains essential even as the optimization boundary expands outward.}
\label{fig:boundary}
\end{figure*}

Figure~\ref{fig:boundary} summarizes the central historical pattern. The vertical axis is the \emph{scope of decision authority}, not benchmark quality. The broad movement is from execution inside one replica toward control over pools, state, heterogeneous capacity, and workload classes.

\subsection{Scheduling becomes continuous}
Autoregressive generation breaks assumptions inherited from static DNN inference. Requests have different lengths, and a request that finishes should not force later arrivals to wait for the rest of its batch. Orca's iteration-level scheduling and selective batching moved the scheduling granularity from request to generation iteration \citep{yu2022orca}. The important idea is broader than its 36.9$\times$ reported comparison: the scheduler must observe progress while execution is ongoing.

vLLM built on continuous batching but exposed a different limiting resource: KV-cache memory. PagedAttention treats KV pages like virtual-memory pages, reducing fragmentation and enabling sharing across sequences \citep{kwon2023pagedattention}. The practical result is not merely memory savings. Better KV utilization increases the number of concurrent sequences the engine can batch, which changes throughput and latency simultaneously. In that sense, vLLM made \emph{state allocation} part of the scheduling problem.

\subsection{The engine becomes hardware-aware}
Attention and matrix kernels remain fundamental. FlashAttention-3 uses Hopper-specific asynchrony, warp specialization, and low precision, reporting 1.5--2.0$\times$ speedup over FlashAttention-2 on H100, up to 740 TFLOP/s in FP16 and near 1.2 PFLOP/s in FP8 \citep{shah2024flashattention3}. MInference exploits dynamic sparse attention for long-context prefill and reports up to 10$\times$ prefill acceleration on A100 while maintaining accuracy in its evaluated tasks \citep{jiang2024minference}. Mnemosyne combines adaptive chunking and multiple parallelism dimensions to support interactive inference at context lengths up to 10 million tokens under its reported TBT target \citep{agrawal2024mnemosyne}.

These examples matter to the control-plane thesis because the controller can only make useful decisions if endpoint capabilities are characterized correctly. ``GPU count'' is not a sufficient capacity model when kernels, quantization, parallelism, model architecture, context length, and phase all change service rates.

\subsection{Prefill and decode stop looking like one job}
Sarathi-Serve shows that full prefill/decode separation is not the only way to reduce phase interference. Its chunked-prefill scheduler splits long prefills into smaller chunks so decode batches are not stalled by monolithic prompt processing \citep{agrawal2024sarathi}. Splitwise and DistServe instead disaggregate the phases into different resource pools \citep{patel2024splitwise,zhong2024distserve}. Splitwise reported 1.4$\times$ higher throughput at 20\% lower cost in one cluster design, or 2.35$\times$ throughput at the same cost and power in another \citep{patel2024splitwise}. DistServe reported that, across its evaluated settings, disaggregation and phase-specific parallelism could serve 7.4$\times$ more requests or meet 12.6$\times$ tighter SLOs while keeping more than 90\% of requests within latency constraints \citep{zhong2024distserve}.

A necessary caveat is that \emph{disaggregation itself is not a universal throughput optimization}. Current vLLM documentation explicitly notes that disaggregated prefill does not inherently improve throughput; its direct value is separate TTFT/ITL control and tail-latency isolation \citep{vllm2026disagg}. Throughput gains reported by larger distributed stacks arise from the complete configuration--pool sizing, batching, routing, parallelism, networking, cache reuse, and workload shape--not from moving prefill and decode apart by definition. This distinction is important for both research claims and operator guidance.

\section{KV State Becomes a Distributed Resource}
The KV cache began as engine-local memory, but repeated prompts, long sessions, and phase disaggregation push its useful lifetime beyond a single request and often beyond a single GPU.

Preble makes this shift explicit. It co-optimizes distributed prompt sharing and load balance, reporting 1.5--14.5$\times$ lower average latency and 2--10$\times$ lower p99 latency than its compared systems across evaluated workloads \citep{srivatsa2024preble}. Mooncake goes further by organizing Kimi serving around a disaggregated KV cache that uses GPU memory together with otherwise underutilized CPU, DRAM, and SSD resources. It reports up to 525\% throughput increase in selected simulated long-context scenarios and 75\% more requests handled under its real workload evaluation \citep{qin2024mooncake}. MemServe similarly proposes an elastic distributed memory pool and global prompt-tree scheduling to combine context caching with disaggregated inference \citep{hu2024memserve}.

The architectural consequence is that routing and state placement can no longer be treated independently. Suppose a prefix of size $S$ tokens resides on endpoint $i$ but endpoint $j$ has the lowest queue. The system has at least four choices: route to $i$, wait for $i$, transfer state from $i$ to $j$, or recompute on $j$. A simple proposed crossover condition is:
\begin{equation}
\begin{split}
T_{transfer}(S,i,j)+T_{queue}(j) \\ < \min\{T_{queue}(i),\;T_{recompute}(S,j)\}.
\end{split}
\label{eq:transfer}
\end{equation}
The same inequality can be expressed in cost or energy rather than time. The important point is that cache locality is not a binary routing rule. It is one term in a dynamic decision whose value depends on queue state, network bandwidth, cache tier, and expected future reuse.

Llumnix provides a related lesson: request state can be migrated while work is running. It reports order-of-magnitude tail-latency improvements, up to 1.5$\times$ acceleration for high-priority requests, and up to 36\% cost savings at similar tail latency in its evaluated scenarios \citep{sun2024llumnix}. Once state is movable, ``where a request started'' need not determine ``where it finishes.''

\section{vLLM and llm-d: Complementary, Not Competing Layers}
\begin{figure*}[t]
\centering
\includegraphics[width=0.96\textwidth]{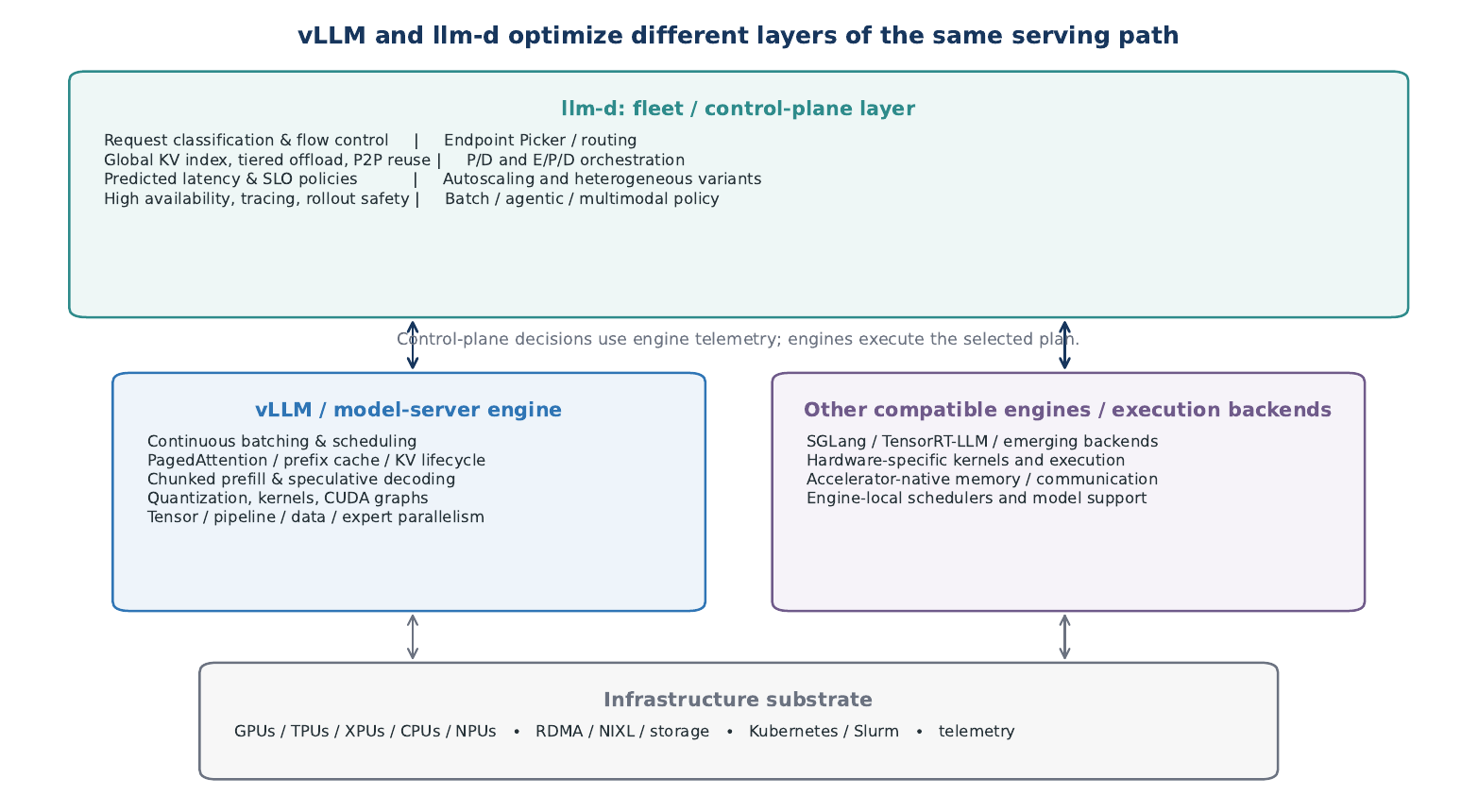}
\caption{A practical layering view. vLLM and peer model servers optimize execution mechanisms; llm-d composes routing, state, topology, scaling, fairness, and reliability decisions across engines and endpoints. The layers exchange telemetry and control.}
\label{fig:layers}
\end{figure*}

Figure~\ref{fig:layers} captures the distinction used throughout this paper. It is intentionally not a strict boundary: modern vLLM includes multi-instance mechanisms such as disaggregated prefill connectors, and llm-d depends on model-server telemetry and capabilities. The difference is one of \emph{primary optimization scope}.

As of the September 2026 software snapshot, vLLM's public project describes a broad engine surface: PagedAttention, continuous batching, chunked prefill, prefix caching, speculative decoding, multiple quantization formats, optimized attention and MoE kernels, tensor/pipeline/data/expert/context parallelism, and disaggregated prefill/decode/encode support across a growing set of accelerators \citep{vllm2026repo}. The latest project release at the time of this review is v0.29.0 \citep{vllm2026v029}. These mechanisms are the foundation on which higher-level orchestration operates.

llm-d's current repository states the complementary goal directly: model servers such as vLLM and SGLang efficiently run models on accelerators, while llm-d provides orchestration and optimizations above them for high-scale real-world traffic \citep{llmd2026repo}. The v0.8 release made that architectural identity explicit by adopting upstream vLLM images for most paths and describing llm-d as an \emph{inference control plane, not a fork of the engine underneath it} \citep{llmd2026v08}. v0.9 hardened that control plane with router high availability, bounded flow-control defaults, a plugin lifecycle, safer rolling updates for disaggregated sets, GPU-utilization-aware routing, broader multimodal routing, KEDA-based autoscaling foundations, and end-to-end tracing \citep{llmd2026v09}.

This separation has two benefits. First, engine innovation can continue independently: a new vLLM scheduler, kernel, quantization method, or accelerator backend becomes an improved execution primitive. Second, the control plane can compare endpoints and topologies without assuming one engine implementation. The v0.8 release's addition of first-class SGLang paths and filesystem-based endpoint discovery for non-Kubernetes environments is evidence that this abstraction is becoming intentional rather than incidental \citep{llmd2026v08}.

\section{The Control Plane: Routing, Flow, Scaling, and Evidence}
The central operational problem is no longer simply balancing request counts. Two requests with the same token length can have different costs depending on cache overlap; two endpoints with the same queue depth can have different accelerators, batch states, and cache pressure; two tenants can have different priorities; and two replicas can be in different rollout or failure states.

\subsection{Routing as state-aware prediction}
A useful progression can be seen inside llm-d itself. Prefix-aware routing uses cache locality to avoid redundant prefill. Token-aware routing then adds a saturation valve: stay with the cache-warm endpoint while it remains within a calibrated load range, otherwise redistribute to prevent affinity from becoming a queue \citep{mitra2026tokenaware}. In the project's three-workload study, the matched configuration sustained 46k versus 16k input tokens/s at the code-generation knee (2.9$\times$), approximately parity on a decode-bound reasoning workload, and 90k versus 45k input tokens/s on the B2B SaaS workload (2.0$\times$) relative to Kubernetes round-robin at their defined knees \citep{mitra2026tokenaware}. The fact that reasoning was near parity is as important as the wins: a routing signal must match the workload bottleneck.

Predicted-latency routing moves another step away from hand-tuned weights. The llm-d/Google implementation trains an online model from features such as prompt length, cache hit, running requests, queue depth, and KV utilization. Its project report states a 43\% reduction in P50 end-to-end latency and 70\% TTFT improvement on a representative MaaS benchmark, and up to 40\% TTFT/ITL reduction in production Vertex AI clusters \citep{mitra2026predicted}. This is evidence that routing can become a learned control problem, though generalization, concept drift, exploration safety, and prediction overhead remain open questions.

\subsection{Flow control and fairness}
Routing after saturation is too late if every backend already holds a long queue. llm-d's flow-control layer therefore queues centrally and admits requests according to fairness and priority policies before they commit to backend work \citep{llmd2026v08,llmd2026v09}. That turns admission into part of inference scheduling. It also introduces an important separation of concerns: the router chooses a feasible execution destination; flow control decides when traffic should enter the serving system and how tenants share scarcity.

This matters for SLOs because waiting in a centralized queue can be more controllable than waiting inside a model server. It also matters for cache reuse: a request held centrally can still be placed using the most recent fleet state rather than being trapped behind earlier work at a single replica.

\subsection{Autoscaling becomes workload-variant scaling}
Traditional horizontal pod autoscaling treats replicas as interchangeable. Distributed LLM serving increasingly violates that assumption. A deployment may contain prefill and decode pools at different tensor-parallel widths, different accelerator types, or different model variants. llm-d's Workload Variant Autoscaler (WVA) direction and KEDA integration reflect a move toward scaling the \emph{shape} of capacity rather than only its replica count \citep{llmd2026v09}.

A controller can therefore ask a richer question: should it add one decode replica, two prefill replicas, a cheaper batch pool, a different hardware variant, or no capacity at all because the bottleneck is cache or network? That is a planning problem rather than a threshold problem.

\subsection{Tracing closes the learning loop}
A learned controller is only as good as the evidence it receives. llm-d's 2026 tracing work propagates OpenTelemetry context across the Gateway, Endpoint Picker, KV-cache path, prefill/decode proxy, and model server, adding decision-level spans rather than only aggregate metrics \citep{liu2026tracing}. This is significant because distributed inference failures are often causal-chain failures: a slow request may be caused by a cache miss that changed routing, a transfer that changed TTFT, or a flow-control decision that changed queueing. Aggregate P95 latency cannot reconstruct that history.

\section{What the Reported llm-d Evidence Actually Shows}
\label{sec:evidence}
\begin{figure*}[t]
\centering
\includegraphics[width=0.96\textwidth]{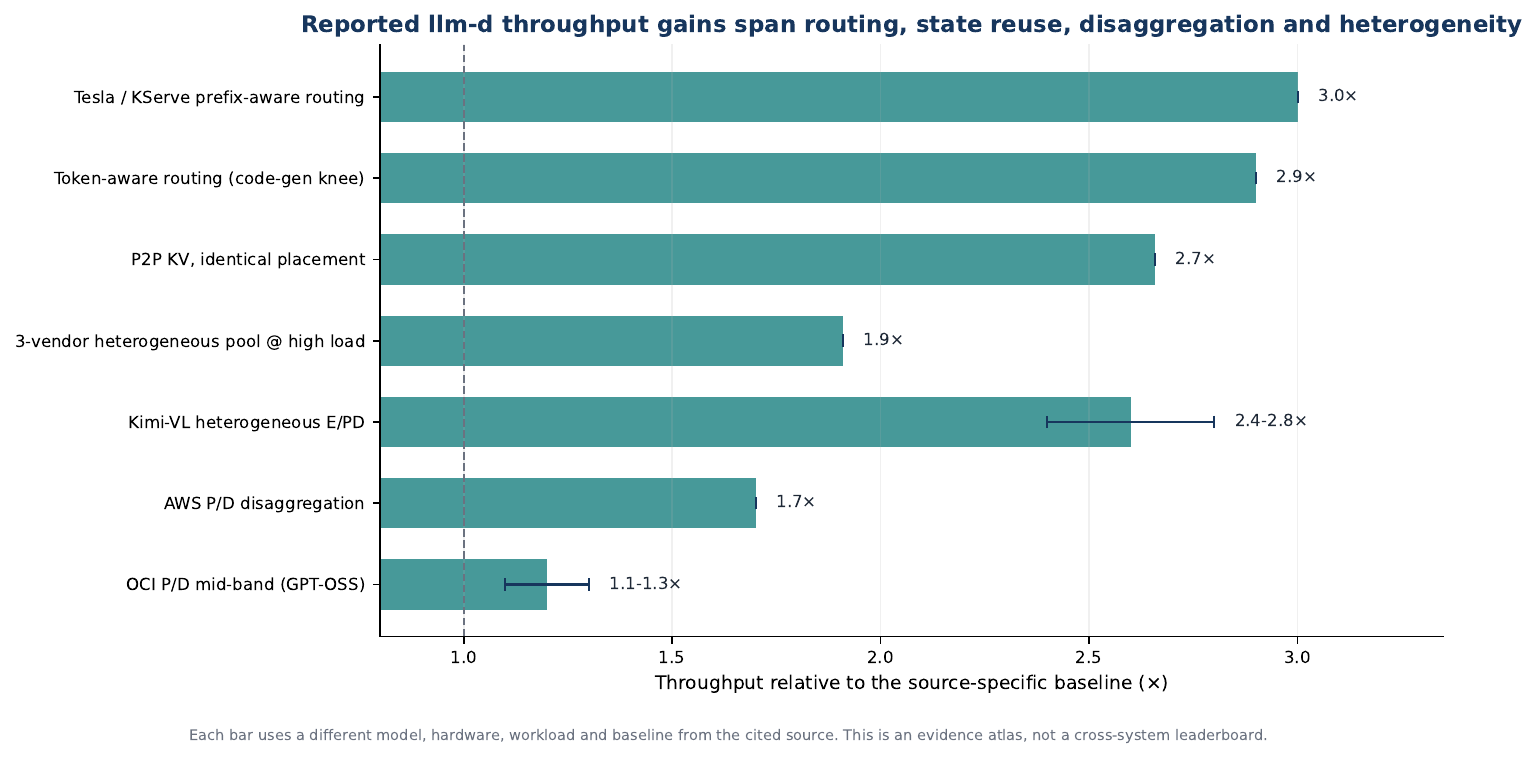}
\caption{Reported llm-d throughput ratios relative to each source's own baseline. Models, hardware, workloads, software versions, and baseline definitions differ. The figure is an evidence atlas, not a leaderboard or meta-analysis.}
\label{fig:throughput}
\end{figure*}

\begin{figure*}[t]
\centering
\includegraphics[width=0.96\textwidth]{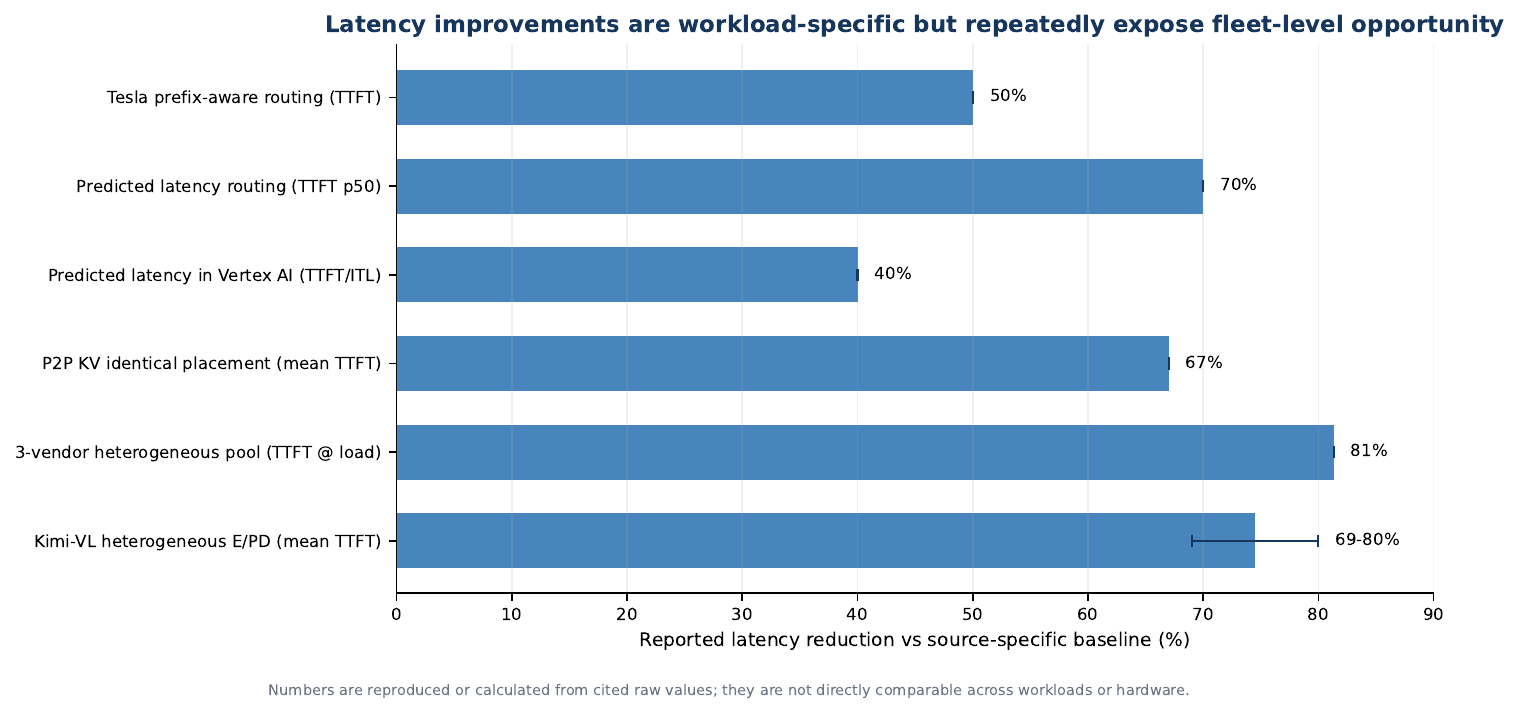}
\caption{Reported latency reductions from source-specific llm-d studies. The 3-vendor value is calculated from the published 36.4 s versus 6.8 s TTFT values; all other plotted values are directly reported ranges or reductions. These results are not directly comparable across workloads.}
\label{fig:latency}
\end{figure*}

The current llm-d evidence is unusually useful because it spans several mechanisms and hardware environments. It must, however, be interpreted as a set of \emph{source-local case studies}. Figures~\ref{fig:throughput} and~\ref{fig:latency} deliberately preserve that limitation.

\begin{table*}[t]
\centering
\scriptsize
\caption{Selected quantitative evidence used in this synthesis. Every row retains its source-specific model, hardware, workload, and baseline; values are not a common benchmark.}
\label{tab:quantitative}
\begin{tabularx}{\textwidth}{p{0.18\textwidth} p{0.25\textwidth} p{0.27\textwidth} X}
\toprule
\textbf{System / mechanism} & \textbf{Evaluation setting (abridged)} & \textbf{Reported result} & \textbf{Interpretation used here} \\
\midrule
vLLM / PagedAttention \citep{kwon2023pagedattention} & SOSP 2023 serving evaluation vs FasterTransformer/Orca & 2--4$\times$ throughput at similar latency in evaluated settings & Efficient KV memory changes achievable batching and engine goodput. \\
Splitwise \citep{patel2024splitwise} & Phase-split cluster designs & 1.4$\times$ throughput at 20\% lower cost in one design; 2.35$\times$ throughput at same cost/power in another & Prefill and decode can justify different hardware and resource policies. \\
DistServe \citep{zhong2024distserve} & Phase-specific SLO-constrained serving & Up to 7.4$\times$ more requests or 12.6$\times$ tighter SLOs while $>$90\% meet constraints & Goodput under phase-specific SLOs is more useful than raw throughput alone. \\
Prefix-aware llm-d routing \citep{tang2026tesla,llmd2026repo} & Llama 3.1 70B, 4 AMD MI300X, project/production report & 3$\times$ higher output throughput and 2$\times$ faster TTFT vs round-robin & Real traffic can make locality-aware placement materially different from request-count balancing. \\
Predicted-latency routing \citep{mitra2026predicted} & Representative MaaS benchmark plus production Vertex AI observations & 43\% lower P50 E2E and 70\% TTFT improvement in benchmark; up to 40\% TTFT/ITL reduction in Vertex & Online predictors can outperform fixed heuristic weights when features reflect queue/cache state. \\
P2P KV sharing \citep{guy2026p2p} & GLM-5.2, identical placement comparison & Mean TTFT 7.85 s $\rightarrow$ 2.56 s; 3.80 $\rightarrow$ 10.10 req/s & State movement can decouple the best execution endpoint from the cache owner. \\
Heterogeneous llm-d pool \citep{kannan2026heterogeneous} & 20-pod, three-vendor Granite-4.1-8B pool & Peak 14.2k vs 9.6k output tok/s; at highest load TTFT 6.8 s vs 36.4 s & Heterogeneous capacity needs capability-aware routing; control-plane CPU can become a bottleneck. \\
AWS P/D study \citep{gangasani2026aws} & GPT-OSS, NVIDIA B200, 1024 input/1024 output & Up to 70\% more tokens/s than the study's standard vLLM baseline & End-to-end disaggregated configurations can improve throughput for some load/topology regimes. \\
vLLM x Mooncake \citep{qiao2026mooncakevllm} & Agentic traces on 12 GB200 GPUs & Cache hit 1.7\% $\rightarrow$ 92.2\%; 3.8$\times$ throughput; 46$\times$ lower P50 TTFT & Long-lived reusable state can dominate agentic inference economics and latency. \\
RaidServe \citep{xu2026raidserve} & 8$\times$H100 fault scenarios & Up to 2$\times$ throughput and two orders of magnitude faster recovery vs compared fault-handling methods & Reliability is part of serving efficiency, not only an operational afterthought. \\
\bottomrule
\end{tabularx}
\end{table*}

\subsection{Prefix-aware routing in production infrastructure}
A Tesla/Red Hat/KServe report describes a production journey from a straightforward vLLM StatefulSet toward KServe + llm-d + vLLM. The llm-d project summarizes a benchmark from that work as approximately 3$\times$ higher output throughput and 2$\times$ faster TTFT with prefix-cache-aware routing versus round-robin on Llama 3.1 70B using four AMD MI300X devices \citep{tang2026tesla}. Because this is a production/project report rather than an independently controlled academic experiment, the correct conclusion is not that prefix routing always yields 3$\times$. It is that real production traffic can contain enough reusable state for load balancing to become a cache-management problem.

\subsection{Peer-to-peer KV: separating placement from reuse}
The P2P KV work is architecturally important because it removes a false choice. Without P2P, cache reuse often implies routing back to the cache owner. With peer transfer, the scheduler can choose a less-loaded endpoint and move state instead. In one GLM-5.2 experiment with identical placement, the project reports mean TTFT falling from 7.85 s to 2.56 s and throughput increasing from 3.80 to 10.10 requests/s (2.7$\times$) when P2P reuse is enabled \citep{guy2026p2p}. A separate Llama-3.1-8B pool reports a 22\% increase in fleet ceiling and 32\% peak token-throughput increase near saturation \citep{guy2026p2p}. The same report also shows a case where local affinity has better median TTFT but load-aware placement plus P2P improves p99 TTFT and throughput. That is exactly the kind of non-monotonic tradeoff a future planner must represent.

\subsection{Heterogeneous pools: endpoint count is not capacity}
In a 20-pod three-vendor Granite-4.1-8B benchmark, llm-d reports a 14.2k output-token/s peak versus 9.6k for Kubernetes round-robin, and at the highest tested load reports 6.8 s TTFT versus 36.4 s with round-robin, with 91\% more throughput at that load \citep{kannan2026heterogeneous}. The report also states that the experiment terminated because the single Endpoint Picker replica became CPU-bound, not because the accelerator pool itself saturated. This is a useful systems result: control-plane compute can become the new bottleneck after accelerator placement improves.

\subsection{Heterogeneous stages in multimodal serving}
A 2026 Kimi-VL study separates the vision encoder from language prefill/decode, placing encoding on four Intel Arc Pro B60 devices and language execution on one NVIDIA H200. Relative to the collocated baseline, the report measures 2.4--2.8$\times$ higher throughput and roughly 69--80\% lower mean TTFT under load \citep{zhang2026kimivl}. This is not a vendor ranking; the paper itself notes that its acquisition-cost sensitivity check excludes host, networking, power, cooling, software, and utilization. The more general conclusion is that \emph{stage heterogeneity can be a first-class placement variable}.

\subsection{Cloud-provider P/D studies}
AWS reports up to 70\% more tokens/s for an llm-d P/D configuration than its standard vLLM baseline on a GPT-OSS 1024-input/1024-output workload on B200 infrastructure as concurrency rises to 128 \citep{gangasani2026aws}. OCI reports workload-dependent throughput-per-GPU gains from disaggregated serving on MI300X infrastructure and emphasizes the middle ground of enterprise concurrency and latency consistency rather than maximum single-user speed \citep{kennetz2026oci}. These cloud reports support deployment feasibility, but they should not be read as contradicting vLLM's statement that disaggregation alone does not improve throughput. The complete systems differ in pool ratios, routing, batching, networking, and workload.

\section{Bottleneck Migration: From FLOPs to State and Decisions}
\label{sec:benchmarks}
Connecting the literature produces a useful model of \emph{bottleneck migration}. An optimization does not remove scarcity; it often moves the scarce resource somewhere else.

\begin{table*}[t]
\centering
\scriptsize
\caption{Bottleneck migration across the modern inference stack. The table connects prior work; it is not a claim that each transition is historically exclusive.}
\label{tab:bottleneck}
\begin{tabularx}{\textwidth}{p{0.13\textwidth} p{0.20\textwidth} X X}
\toprule
\textbf{Bottleneck} & \textbf{Representative mechanisms} & \textbf{What improves} & \textbf{What may become limiting next} \\
\midrule
Autoregressive scheduling & Orca; continuous batching & GPU occupancy; request interleaving & KV memory and batch-state overhead \\
KV memory & PagedAttention; prefix caching & effective batch size; reuse & distributed locality and cache ownership \\
Prefill interference & chunked prefill; P/D separation & TTFT/ITL isolation & transfer bandwidth; pool ratio; queue coordination \\
Distributed locality & Preble; Mooncake; MemServe & avoided recomputation; long-context reuse & hot-cache queues; transfer decision; storage tiers \\
Endpoint selection & cache/load-aware routing; predicted latency & fleet goodput and tail latency & router CPU; stale telemetry; prediction drift \\
Hardware diversity & stage/variant-aware placement & capacity use; phase fit & capability modeling; cross-vendor transfer \\
Reliability & HA routing; RaidServe; tracing & failure tolerance; diagnosis & redundant state cost; recovery policy \\
Composite workloads & E/P/D; M* dataflow graphs & workload-specific placement & general graph scheduling and cross-stage SLOs \\
\bottomrule
\end{tabularx}
\end{table*}

The key implication is that raw FLOPs become a weaker predictor of delivered service as the workload becomes more stateful. Consider a repeated 200k-token agent turn. If 99\% of the context can be reused, the fastest endpoint without that state may be slower than a nominally weaker endpoint with the right cache. The llm-d GLM-5.2 study provides a striking example: across 219 production Claude Code sessions, the median main-agent request carried 195k input tokens and only 317 output tokens; 96\% of main-agent turns reused at least 90\% of their input, and prefix-aware routing captured roughly 96--97\% of available reuse \citep{ayoub2026agentic}. In a separate paused-conversation run, cached later turns reached first token 2.8$\times$ faster than uncached first turns \citep{ayoub2026agentic}.

The same shift appears in the vLLM-Mooncake integration. On the project's realistic agentic trace evaluation, a distributed KV store raised cache hit rate from 1.7\% to 92.2\%, with 3.8$\times$ higher throughput, 46$\times$ lower P50 TTFT, and 8.6$\times$ lower end-to-end latency in the reported 12-GB200 setup \citep{qiao2026mooncakevllm}. Again, these are source-local numbers. Their architectural meaning is that \emph{session state can dominate execution cost}.

The controller's job therefore becomes one of matching demand to reusable state and capacity while avoiding state-induced hotspots. The most valuable telemetry is no longer only GPU utilization. It includes uncached input tokens in flight, cache residency and tier, expected output length, queue age, transfer bandwidth, request priority, session identity, rollout revision, and failure state.

\section{Reliability Becomes an Inference Optimization}
Performance measurements often assume healthy hardware. Production clusters cannot. A single failed device in a tightly coupled tensor-parallel group can stall an otherwise healthy request and force expensive KV recomputation.

RaidServe addresses this explicitly with cyclic KV placement, hybrid attention, load-aware routing, proactive KV backup, and on-demand weight recovery. On an 8$\times$H100 DGX system it reports up to 2$\times$ higher throughput and two orders of magnitude faster recovery than standard fault-handling methods under its evaluated failure scenarios \citep{xu2026raidserve}. llm-d v0.9 addresses a different reliability layer: multiple Endpoint Picker replicas, graceful drain behavior, production-safe flow-control defaults, plugin maturity gates, and revision-aware routing for disaggregated rolling updates \citep{llmd2026v09}.

The connection is that reliability policy consumes the same resources as performance policy. Replicated KV state uses memory and network bandwidth; conservative rollout gating may reduce instantaneous capacity; standby routers and redundant components cost CPU. Therefore reliability should enter the same planning objective as latency and cost rather than being treated as a post-deployment checklist.

One simple proposed metric is \emph{availability-adjusted goodput}:
\begin{equation}
G_{avail} = \frac{N_{SLO\text{-}satisfied\ requests}}{T}\,(1-P_{service\ interruption}),
\end{equation}
where the interruption probability is measured under a declared failure model. The formula is intentionally simple; its purpose is to prevent a system that is very fast when healthy but fragile under expected failures from appearing unambiguously superior.

\section{Agentic and Multimodal Workloads Change the Unit of Scheduling}
Text chat encouraged a request-centric abstraction: prompt in, tokens out. New workloads make that abstraction leaky.

\subsection{Agent programs are sessions, not isolated prompts}
Agentic workflows alternate model calls with tool execution, retrieval, compilers, terminals, and other external actions. A session may pause for seconds or minutes and then return with nearly the same long prefix plus new observations. The llm-d GLM-5.2 trace analysis shows a workload where input accounts for 98.6\% of served tokens and reusable prefix state dominates deployment design \citep{ayoub2026agentic}. The vLLM-Mooncake results similarly show that distributed KV persistence can change performance by orders of magnitude for such trace shapes \citep{qiao2026mooncakevllm}.

This suggests that the scheduling entity should sometimes be a \emph{session} or \emph{agent program}, with policy for state retention during tool-call pauses, rather than a sequence of independent HTTP requests. The control plane needs to decide whether to pin a session, offload it, transfer it, or evict it based on reuse probability and pause duration.

\subsection{Multimodal models are graphs, not just longer prompts}
Multimodal serving adds a different kind of heterogeneity. llm-d's multimodal analysis notes that image/video content changes request size, cache identity, stage structure, and token-cost estimation; text-tuned routing assumptions can remain syntactically valid while becoming semantically wrong \citep{yu2026multimodal}. The Kimi-VL heterogeneous E/PD results show one practical stage split \citep{zhang2026kimivl}.

Stanford/UW/CMU's M* generalizes this idea. It represents composite models as dataflow ``Walk Graphs'' whose components can be placed and optimized independently. On Qwen3-Omni TTS, M* reports up to 2.7$\times$ higher throughput than vLLM-Omni and 4$\times$ higher throughput than SGLang-Omni in the project-reported configuration; on BAGEL text-to-image it reports 20\% lower average end-to-end latency than vLLM-Omni \citep{jha2026mstar}. The conceptual implication is larger than those numbers: prefill/decode may be the first useful instance of a more general \emph{inference-graph scheduling} problem.

\section{Independent Convergence: NVIDIA Dynamo and Cluster Control}
NVIDIA Dynamo provides useful external confirmation of the engine/control-plane split. Its current documentation describes an open-source inference framework that works with vLLM, SGLang, and TensorRT-LLM; it provides disaggregated serving, KV-aware routing, cache management, and autoscaling around model engines \citep{nvidia2026dynamo}. Its KV-aware router explicitly combines reusable KV state with projected active load, and its disaggregated path separates worker selection from KV transfer \citep{nvidia2026dynamo}.

This architectural convergence matters because it reduces the risk that the inference-control-plane idea is a project-specific abstraction. Different communities are arriving at similar primitives: engine interoperability, cache-aware selection, independent phase pools, topology-aware transfer, autoscaling, and policy layers. The research opportunity is therefore not to argue over which project owns the abstraction. It is to formalize the decision problem and develop reproducible ways to compare policies across engines and hardware.

DynamoLLM provides a parallel research signal from Microsoft. It dynamically reconfigures the number of instances, model parallelism, and GPU frequency to meet performance SLOs while optimizing energy and cost, reporting 53\% energy, 38\% operational-carbon, and 61\% customer-cost reductions in its evaluated service-level scenarios \citep{stojkovic2025dynamollm}. CMU's inference-energy study reaches a related conclusion from a different direction: optimization effectiveness is highly sensitive to workload geometry, software stack, hardware, decoding strategy, and parallelism, with properly selected optimizations reducing energy by up to 73\% versus its unoptimized baseline \citep{fernandez2025energy}. Together they argue for workload-aware planning rather than a static ``green'' configuration.

\section{Proposed Architecture: The Inference Execution Planner}
\begin{figure*}[t]
\centering
\includegraphics[width=0.96\textwidth]{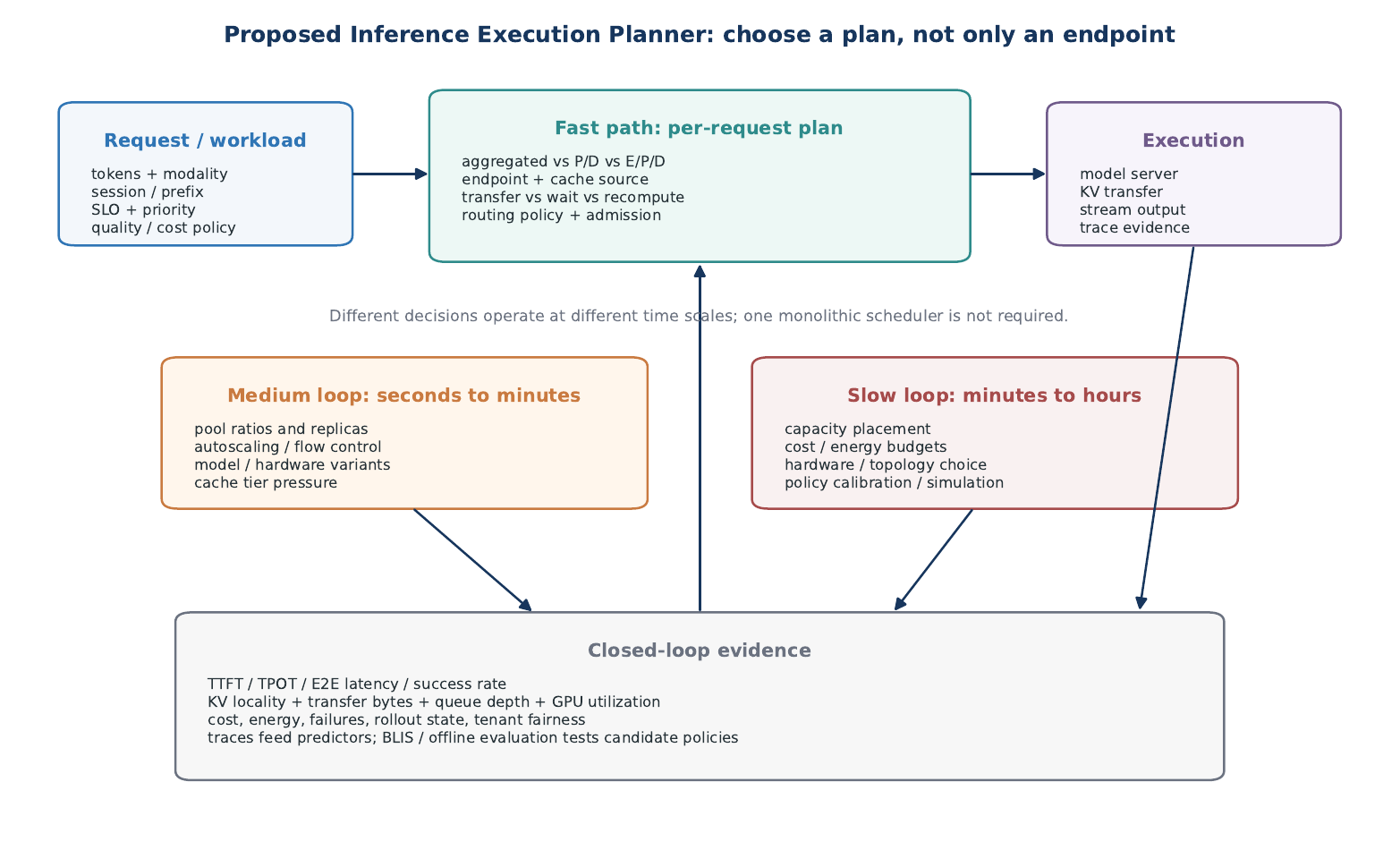}
\caption{Proposed Inference Execution Planner. Fast decisions choose a per-request execution plan; medium and slow loops reshape the feasible plan set by changing pool ratios, replicas, hardware variants, cache tiers, and capacity. Telemetry and simulation close the loop. This architecture is a proposal, not a current llm-d feature claim.}
\label{fig:planner}
\end{figure*}

The preceding evidence suggests that selecting an endpoint is becoming too narrow a formulation. A request may need a plan that includes topology, state movement, hardware, admission, and future capacity implications. Figure~\ref{fig:planner} presents a proposed Inference Execution Planner (IEP).

\subsection{Plan space}
For request or session $r$, let a candidate execution plan $p$ specify:
\begin{equation}
p = (m,h,g,k,x,a,s),
\end{equation}
where $m$ is the model/model variant, $h$ the hardware placement, $g$ the execution graph (aggregated, P/D, E/P/D, or a general stage graph), $k$ the KV-state source/tier/action, $x$ the transfer and network path, $a$ the admission/routing policy, and $s$ the scaling/replica context. Not every request can choose every dimension at millisecond latency; the planner is intentionally hierarchical.

The fast path chooses among currently available plans. The medium loop changes the pool ratios and replicas over seconds to minutes. The slow loop handles capacity placement, model/hardware variants, budget policy, and offline simulation over minutes to hours. This timescale decomposition avoids building one monolithic controller with impossible latency and observability requirements.

\subsection{Constraint-first selection}
A practical planner should reject infeasible plans before applying a weighted objective. Let $\mathcal{P}_r$ be the candidate plans and $\mathcal{F}_r$ those satisfying hard constraints:
\begin{align}
\mathcal{F}_r = \{p \in \mathcal{P}_r : &\ \widehat{TTFT}_{r,p} \leq S_{TTFT},\\
&\ \widehat{TPOT}_{r,p} \leq S_{TPOT},\\
&\ Q_{r,p}\geq Q_{min},\ A_{p}\geq A_{min},\\
&\ Policy(r,p)=1 \}.
\end{align}
Among feasible plans, an operator could optimize a policy-specific score:
\begin{equation}
p^* = \arg\min_{p\in \mathcal{F}_r}
\left[\alpha \widehat{L}_{r,p}+\beta \widehat{C}_{r,p}+\gamma \widehat{E}_{r,p}+\delta \widehat{R}_{r,p}\right],
\label{eq:objective}
\end{equation}
where $L$ is latency, $C$ monetary serving cost, $E$ energy, and $R$ reliability/risk cost. The terms are not assumed commensurate; in practice a lexicographic or Pareto policy may be safer than one opaque scalar. The point is that the control plane can optimize multiple operational dimensions while preserving declared SLO and quality guardrails.

\subsection{State action is part of the plan}
The IEP should expose state actions explicitly: local hit, owner routing, peer transfer, tier restore, recompute, retain, or evict. Current P2P work demonstrates why this matters: the best execution endpoint and the cache owner need not be the same \citep{guy2026p2p}. For agentic sessions, expected future reuse can justify keeping state in a cheaper tier even when it is not needed immediately. For multimodal graphs, different stages can have different cache identities and lifetimes.

\subsection{Simulation before accelerator experiments}
Searching this plan space entirely on production GPUs is costly. BLIS, an IBM/llm-d discrete-event simulator, reports median 7--9\% error on end-to-end and inter-token latency across 36 validation experiments and approximately 200$\times$ faster evaluation than equivalent cluster runs \citep{toslali2026blis}. Its admission-control case study reports a simulator-guided policy that reduced critical-tier TTFT p90 by up to 97\% and end-to-end latency by up to 50\%, followed by real-cluster validation \citep{toslali2026blis}. Simulation does not replace measurement, but it can be the inner loop for pruning poor plans before scarce accelerator time is used.

\section{Research Questions for the Next Generation of Inference}
The IEP proposal is useful only if it creates testable questions. The following research agenda follows directly from gaps exposed by the literature and current project evidence.

\subsection{RQ1: When should a request be aggregated, P/D, E/P/D, or graph-scheduled?}
The choice should depend on input/output geometry, modality, model architecture, network transfer, queue state, hardware characteristics, and SLO. The vLLM caveat that disaggregated prefill does not automatically increase throughput is a useful baseline \citep{vllm2026disagg}. A future controller should predict the crossover, not encode one topology as universally superior.

\subsection{RQ2: When should state move, work move, or neither?}
P2P KV, Llumnix request migration, tiered caches, and distributed stores all trade data movement against recomputation and queueing \citep{guy2026p2p,sun2024llumnix,qin2024mooncake}. A unified policy should model network contention and the future value of retained state, not just immediate TTFT.

\subsection{RQ3: Can latency prediction become execution-plan prediction?}
Predicted-latency routing shows that live traffic can train useful endpoint-level models \citep{mitra2026predicted}. The next step is predicting a vector: TTFT, TPOT, cost, energy, transfer time, and failure risk for a plan. Research is needed on online calibration, uncertainty bounds, cold-start behavior, and safe fallback when the predictor drifts.

\subsection{RQ4: How should routing and autoscaling be co-designed?}
Routing changes cache locality and queue distribution; autoscaling changes the candidate set and initially creates cold caches. Treating them as independent loops can create oscillation. A joint controller should reason about warm-up, cache replication, draining, and the value of preserving state before adding or removing replicas.

\subsection{RQ5: What is the correct abstraction for heterogeneous capacity?}
The three-vendor and Kimi-VL experiments show that nominal replica counts hide capability differences \citep{kannan2026heterogeneous,zhang2026kimivl}. A capacity unit may need to be workload-specific: tokens/s at a declared SLO for a particular phase and model. That implies capability discovery APIs and continuous calibration rather than static accelerator labels.

\subsection{RQ6: How should reliability enter the planner?}
RaidServe and llm-d v0.9 demonstrate two levels of fault tolerance \citep{xu2026raidserve,llmd2026v09}. Open questions include how much KV redundancy is worth storing, when to proactively replicate state, how to route around degraded tensor-parallel groups, and how to include rollout/failure state in learned policies without making the model brittle.

\subsection{RQ7: Does P/D become general graph scheduling?}
M* and multimodal E/P/D serving suggest a move from two phases to arbitrary component graphs \citep{jha2026mstar,yu2026multimodal}. A future planner may need to map graph stages onto hardware, fuse or split stages dynamically, and optimize end-to-end SLOs rather than per-stage throughput.

\subsection{RQ8: How should agent sessions be priced and scheduled?}
Agentic traces make input reuse and pause time central \citep{ayoub2026agentic,qiao2026mooncakevllm}. Research should evaluate session-aware state retention, cache tenancy, fairness across long-lived agents, and admission policies that prevent a few massive sessions from monopolizing reusable-state capacity.

\subsection{RQ9: What should a reproducible control-plane benchmark look like?}
A useful benchmark suite must vary prompt/output geometry, prefix sharing, session pauses, modality, SLO, load regime, failure events, hardware mix, and network topology. It should report both raw throughput and \emph{goodput}: work completed within declared SLOs. It should also publish the routing and cache-hit traces needed to explain why one policy wins.

\section{An Evaluation Framework for Industry and Research}
A central weakness in inference comparisons is that ``tokens per second'' often collapses several distinct questions. A production-oriented evaluation should report at least five dimensions.

\begin{table*}[t]
\centering
\small
\caption{Recommended evaluation dimensions for a distributed inference control plane.}
\label{tab:metrics}
\begin{tabularx}{\textwidth}{p{0.15\textwidth} X X}
\toprule
\textbf{Dimension} & \textbf{Core metrics} & \textbf{Why it matters} \\
\midrule
User latency & TTFT, TPOT/ITL, E2E P50/P95/P99, deadline/SLO hit rate & Separates startup delay, streaming quality, and tail behavior. \\
Useful capacity & successful req/s, input/output tok/s, SLO-goodput, batch completion rate & Avoids rewarding work that violates the service contract. \\
State efficiency & prefix hit rate, reusable-token ratio, transfer bytes, tier hit rate, recompute avoided & Explains whether capacity comes from execution or state reuse. \\
Resource economics & accelerator-hours, cost/request, cost/SLO-goodput, energy/request, network/storage cost & Connects performance to deployability and sustainability. \\
Reliability / operations & failover time, interruption probability, rollback success, queue age, fairness, control-plane CPU & Measures behavior under faults, saturation, and day-two operations. \\
\bottomrule
\end{tabularx}
\end{table*}

For workload $W$, define SLO-goodput as:
\begin{equation}
G_{SLO}(W) = \frac{N_{\mathrm{successful\ and\ SLO\!\!-satisfied}}(W)}{T}.
\end{equation}
The metric can be expressed in completed requests or useful output tokens. For cost-aware comparison, report SLO-goodput per dollar and accelerator-hours per SLO-satisfied request. For stateful workloads, publish cache hit and reusable-token distributions alongside throughput, because a cache-friendly trace can make a routing policy appear universally superior when it is not.

The experimental matrix should include: short chat, long-context/RAG, code/agentic sessions, reasoning-heavy decode, offline batch, multimodal workloads, and at least one failure/rollout scenario. Each workload should sweep offered load from underutilization through the saturation knee. If a system uses learned routing, the evaluation should include warm-up and drift periods rather than training and testing on a stationary trace only.

\section{Practical Guidance for Platform Teams}
The research can be translated into several conservative engineering rules.

\begin{enumerate}
\item \textbf{Optimize the engine first, then the fleet.} Control-plane sophistication cannot compensate for an inefficient model server. Use modern batching, memory management, kernels, quantization, and parallelism before attributing every problem to routing.
\item \textbf{Measure the workload geometry.} Input/output distributions, prefix reuse, session pauses, modality, and concurrency determine which optimization matters. A routing policy tuned for long shared prefixes can be irrelevant for decode-bound reasoning traffic \citep{mitra2026tokenaware}.
\item \textbf{Treat KV as state with an economic value.} Track residency, bytes transferred, recomputation avoided, and cache pressure. ``Cache hit rate'' without the size/value of the hit can be misleading.
\item \textbf{Do not disaggregate by ideology.} Compare aggregated, chunked-prefill, and P/D configurations under the same SLO and workload. Include the network transfer path in TTFT. vLLM's own documentation is a useful reminder that the split alone is not a throughput guarantee \citep{vllm2026disagg}.
\item \textbf{Route on the bottleneck signal.} Prefix affinity, token load, active requests, predicted latency, and GPU utilization are not interchangeable. The matched signal should correspond to the limiting resource.
\item \textbf{Keep admission outside saturated backends.} Central flow control can protect fairness and SLOs better than allowing every endpoint to build an opaque queue.
\item \textbf{Scale variants, not only replicas.} If prefill, decode, multimodal encoders, or hardware classes differ, autoscaling should reason about which type of capacity is missing.
\item \textbf{Make the control plane observable.} Request-level traces should capture routing scores, cache decisions, queueing, transfers, phase boundaries, and the final engine response \citep{liu2026tracing}.
\item \textbf{Test failure and rollout paths as performance scenarios.} A fast system that loses every in-flight session on one GPU or router failure is not production-efficient.
\item \textbf{Use simulation and small-scale experiments before fleet-wide policy search.} Tools such as BLIS illustrate how an offline inner loop can reduce expensive GPU experiments \citep{toslali2026blis}.
\end{enumerate}

\section{What the Future Likely Looks Like}
The evidence supports a direction, not a precise calendar. Three trends appear robust.

First, \textbf{engines will continue to become more capable and more heterogeneous}. vLLM's current feature surface already includes execution paths for dense, MoE, hybrid, multimodal, embedding, and other model classes across many accelerators \citep{vllm2026repo}. The control plane therefore needs capability discovery rather than hard-coded assumptions about a ``GPU replica.''

Second, \textbf{inference state will become increasingly portable}. Distributed KV stores, peer transfer, offload tiers, and request migration all weaken the coupling between a request and one device \citep{qin2024mooncake,hu2024memserve,guy2026p2p,sun2024llumnix}. The limiting question becomes how much state should move, when, and at what cost.

Third, \textbf{the request path will become a policy-driven graph}. Agentic sessions add persistent state; multimodal models add component graphs; P/D and E/P/D expose phase placement; reliability adds replicas and recovery; budget policies add cost constraints. In that environment, the output of the scheduler is naturally an execution plan.

This future is not synonymous with one project. llm-d, NVIDIA Dynamo, vLLM's distributed mechanisms, SGLang, cloud-provider systems, and academic prototypes will continue to overlap and compete. The durable contribution for the research community is to make the decision problem explicit and develop common interfaces, traces, and benchmarks that let different implementations be compared honestly.

\begin{table*}[t]
\centering
\small
\caption{Capability horizon derived from the evidence. These are maturity bands, not calendar promises.}
\label{tab:horizon}
\begin{tabularx}{\textwidth}{p{0.10\textwidth} p{0.19\textwidth} X X}
\toprule
\textbf{Horizon} & \textbf{State of practice} & \textbf{Representative capabilities} & \textbf{Research / engineering frontier} \\
\midrule
H0 & Engine efficiency is established & Continuous batching, PagedAttention, optimized kernels, quantization, prefix caching, TP/PP/EP, speculative decoding & Better per-engine capability discovery and workload-specific performance models. \\
H1 & Fleet intelligence is deployable today & Cache/load-aware routing, global KV indexing, P/D orchestration, flow control, autoscaling, tracing, heterogeneous variants & Joint routing/scaling stability, warm-cache-aware rollout, multi-tenant fairness, control-plane scalability. \\
H2 & Execution-plan control is emerging & P2P state movement, learned latency routing, E/P/D, agent-session retention, simulator-guided policy search & Predict topology + state action + hardware + admission as one constrained plan with uncertainty bounds. \\
H3 & General workload-aware planning is open & Composite multimodal graphs, resilient state, multi-region/cluster choices, policy-driven cost/energy/reliability & Common plan interfaces, graph scheduling, failure-aware optimization, trustworthy online learning, reproducible cross-engine benchmarks. \\
\bottomrule
\end{tabularx}
\end{table*}

\section{Limitations and Responsible Interpretation}
This paper has several limitations. First, it combines peer-reviewed papers with preprints, open-source project reports, and vendor/cloud benchmarks. The evidence hierarchy in Table~\ref{tab:evidence} is intended to keep those categories visible, but it does not eliminate publication or reporting bias. Project blogs naturally highlight successful configurations.

Second, the benchmark atlas is intentionally heterogeneous. Ratios from Tesla, Google, Intel/IBM/Red Hat, AWS, OCI, and llm-d project studies should not be compared as if they were generated from the same harness. Hardware, model, quantization, network, request distribution, concurrency, SLO, and baseline tuning all differ. The figures show that fleet-level choices can matter; they do not establish a universal expected speedup.

Third, the current software landscape changes quickly. This review uses a snapshot through September 13, 2026. llm-d v0.9 and vLLM v0.29 are current reference points, not stable endpoints. APIs and project maturity labels may change after publication.

Fourth, the IEP is a research proposal. It is not claimed to exist as one component in llm-d, vLLM, NVIDIA Dynamo, or any other cited project. Many of its primitives already exist separately, which is precisely why a unifying plan abstraction is worth testing. Whether one planner can outperform carefully composed specialized policies without adding excessive complexity remains an empirical question.

Finally, optimization should not be confused with maximizing machine utilization. The objective must remain tied to useful, reliable service. A system that drives accelerator utilization to 100\% while violating user SLOs, starving tenants, increasing failure blast radius, or spending more to recompute discarded work is not efficient in the operational sense.

\section{Conclusion}
The trajectory of LLM inference is not a story in which one scheduler replaces another. It is a story in which the boundary of what the system is allowed to optimize keeps moving outward.

Orca made autoregressive progress visible to the scheduler. vLLM made KV memory manageable enough to sustain larger continuous batches. Kernel and long-context work improved the execution substrate. Sarathi-Serve, Splitwise, and DistServe exposed phase asymmetry. Preble, Mooncake, MemServe, and Llumnix turned state locality and mobility into distributed decisions. vLLM has continued to absorb stronger execution mechanisms, while llm-d has increasingly organized routing, cache intelligence, phase orchestration, flow control, autoscaling, heterogeneous placement, and reliability above engines. NVIDIA Dynamo's similar architecture suggests that this control-plane direction is broader than one community.

The most useful connecting insight is that modern inference is increasingly constrained by \emph{managed state, placement, and decisions}, not raw FLOPs alone. A request can be expensive because its prefix is cold, because its warm endpoint is saturated, because the KV transfer path is slow, because it was sent to the wrong hardware class, because prefill and decode interfere, or because a failure forces recomputation. Improving the engine remains necessary, but efficient production serving requires coordinating those choices across a fleet.

That motivates the central research proposition of this paper: the next-generation scheduler should be evaluated as an \emph{Inference Execution Planner}. It should choose among feasible execution plans, not merely endpoints, while preserving latency, quality, reliability, fairness, and budget constraints. The immediate research task is not to make that planner maximally autonomous. It is to make its decisions observable, reproducible, failure-aware, and grounded in measurements that can be compared across engines and hardware.

If the field succeeds, the benefit is practical: fewer wasted accelerators, lower latency, more predictable cost, more resilient services, and a serving layer capable of supporting long-context, agentic, multimodal, and future composite models without rebuilding the infrastructure for every new workload. That is a useful direction for both research and industry because it turns inference optimization from a collection of isolated tricks into a measurable systems discipline.

\section*{Acknowledgments and Research Integrity Note}
This manuscript is an independent systems synthesis. No new benchmark measurement, vendor result, or open-source capability is claimed as original experimental evidence. Numerical findings and system capabilities remain attributed to their cited authors, projects, or organizations. The optimization-boundary taxonomy, bottleneck-migration interpretation, Inference Execution Planner proposal, evaluation framework, and research questions are the author's synthesis of those sources.

\bibliographystyle{plainnat}
\bibliography{references}

\end{document}